# Aradeğerleme ve Ardından Geri İteratif Projeksiyon Kullanarak İmge Çözünürlüğü Geliştirme

# Image Resolution Enhancement by Using Interpolation Followed by Iterative Back Projection

Pejman Rasti, Hasan Demirel
Elektrik ve Elektronik Mühendisliği Bölümü
Doğu Akdeniz Üniversitesi
Gazimağusa, Kuzey Kıbrıs Türk Cumhuriyeti
pejman.rasti@cc.emu.edu.tr ve hasan.demirel@emu.edu.tr

Gholamreza Anbarjafari
Elektrik ve Elektronik Mühendisliği Bölümü
Uluslararası Kıbrıs Üniversitesi
Lefkoşa, Kuzey Kıbrıs Türk Cumhuriyeti
sjafari@ciu.edu.tr

*Özetçe*—Bu yazıda, aradeğerleme ve ardından iteratif geri projeksiyon (IGP) çakıştırma yöntemleri kullanarak yeni bir süper çözünürlük tekniği önerilmektedir. Düşük çözünürlüklü giriş imgeleri, eşdeğerleme yapıldıktan sonra IBP yöntemiyle çakıştırılarak yüksek çözünürlüğe sahip daha net çıkış imgeleriüretilmektedir. Önerilen teknik Lena, Elaine, Biber ve Baboon imgeleri üzerinde test edilmiştir. Tepe sinyal-gürültü oranı (TSGO) ve yapısal benzerlik endeksi (YBEN) gibi nicel sonuçları ile görsel sonuçlar konvansiyonel ve literatürde önde gelen süper çözünürlük yöntemleri ile karşılaştırıldığında önerilen tekniğin üstünlüğünü göstermektedir. Örneğin, Lena imgesi için, 6.52 dB daha yüksek bir PSNR değeri elde edilebilmektedir.

*Anahtar Kelimeler* — Süper Çözünürlük; İteratif Geri Projeksiyon; İmge Çakıştırma.

*Abstract*—In this paper, we propose a new super resolution technique based on the interpolation followed by registering them using iterative back projection (IBP). Low resolution images are being interpolated and then the interpolated images are being registered in order to generate a sharper high resolution image. The proposed technique has been tested on Lena, Elaine, Pepper, and Baboon. The quantitative peak signal-to-noise ratio (PSNR) and structural similarity index (SSIM) results as well as the visual results show the superiority of the proposed technique over the conventional and state-of-art image super resolution techniques. For Lena's image, the PSNR is 6.52 dB higher than the bicubic interpolation.

## I. Introduction

The spatial resolution is an important parameter for any digital image system and it refers to the pixel density in an image. Nowadays there is a big demand on resolution enhancement in digital imaging systems. The desire for high-resolution comes from two principal application areas [1]:

i. Improvement of pictorial information for human interpretation
ii. Helping representation for automatic machine perception

The image spatial resolution is first limited by the imaging sensors or the imaging acquisition device. The sensor size or equivalently the number of sensor elements per unit area in the first place determines the spatial resolution of the image to capture. Hence, In order to increase the spatial resolution of an imaging system, one straightforward way is to increase the sensor density by reducing the sensor size. However, as the sensor size decreases, the amount of light incident on each sensor also decreases, causing the so-called shot noise. Also, the hardware cost of a sensor increases with the increase of sensor density or corresponding image pixel density.

Another approach for enhancing the resolution is by employing various signal processing tools. These techniques are specifically referred to as Super-Resolution (SR) reconstruction [2]. High-resolution (HR) images are constrained by low-resolution (LR) images in SR technique.

Huang and Tsay [3] started researching about the SR field in 1984. There was a significant speared in this field after that. Approaches using Frequency Domain [4-6], Bayesian [7], Example-Based [7-11], Set Theoretic [12,13] and Interpolation [14] have been applied to SR techniques.

The application of SR techniques covers a wide range of purposes such as Surveillance video [15-16], remote sensing [17], Medical imaging such as Computerized Tomography (CT) scan, Magnetic Resonance Imaging (MRI), Ultrasound [18-21].

The basic idea behind SR is to combine the non-redundant information contained in multiple LR frames to generate an HR image. A closely related technique with SR is the single-image interpolation approach, which increase the size of the image. However, since the estimated pixel values in interpolation process are generated based on a pre-defined so called interpolation formula which is independent of the nature of the image, e.g. the high and low frequency components of the image, the quality of the single-image interpolation is very much limited due to the ill-posed nature of the problem.

Interpolation is the process of determining the values of a function between its samples. This process accomplishes by conforming to a continuous function through the discrete input samples. The famous Interpolation techniques are nearest neighbor interpolation, bilinear interpolation, and bicubic interpolation.

Nearest neighbor from a computational standpoint is the simplest method of interpolation. Each interpolated output pixel is assigned the value of the nearest sample point in the input image. For nearest neighbor defines the interpolation kernel as

$$h(x) = \begin{cases} 1, & 0 \leq |x| < 0.5 \\ 0, & 0.5 \leq |x| \end{cases} \quad (1)$$

In frequency domain nearest neighbor interpolation is

$$H(w) = \sin c \left(\frac{w}{2}\right) \quad (2)$$

Bilinear passes a straight line through every two consecutive points of the input signal. Bilinear interpolation is corresponding to convolving the sampled input with the following kernel in spatial domain.

$$h(x) = \begin{cases} 1 - |x|, & 0 \leq |x| < 1 \\ 0, & 1 \leq |x| \end{cases} \quad (3)$$

In frequency domain bilinear interpolation is

$$H(w) = \sin c^2 \left(\frac{w}{2}\right) \quad (4)$$

For bicubic interpolation, the block uses the weighted average of four translated pixel values for each output pixel value. For bicubic interpolation defines the interpolation kernel as

$$h(x) = \begin{cases} (a+2)|x|^3 - (a+3)|x|^2 + 1, & 0 \leq |x| < 1 \\ a|x|^3 - 5a|x|^2 + 8a|x| - 4a, & 1 \leq |x| < 2 \\ 0, & 2 \leq |x| \end{cases} \quad (5)$$

In frequency domain bicubic interpolation is

$$H(w) = \frac{12}{w^2}\left(\sin c^2 \left(\frac{w}{2}\right) - \sin c(w)\right) \\ + a\frac{8}{w^2}(\sin c^2(w) - 2\sin c(w) - \sin c(2w)) \quad (6)$$

There are some SR techniques which are using multiple LR images as input in order to generate a super resolved image. These techniques are benefiting from registration. One of the fundamental steps in SR process is the registration of the images [22]. Registration is one of the basic and the principal subjects in the image processing and there are numerous algorithms for it [23]. There are various techniques of registration [24-27] in which here in this research work iterative back projection (IBP) registration [22] is the used.

The proposed technique is using IBP in the registration stage. IBP model has two critical steps, first is to construct the model for imaging process and the second step is image registration. The first step can be described as

$$g_k(y) = DH^{psf} \times f(x) + n_k \quad (7)$$

where $g_k$ are $k^{th}$ observed LR images, y denote the pixel of LR images influenced by the area of x of the SR image f, $H^{psf}$ is the PSF of blur kernel, D means decimating operator and $n_k$ is an additive noise term.

In this method firstly a true SR image is assumed. Based on the imaging model given in eqn. (7), different LR images are evaluated. Given the calculated LR images a new SR image is being obtained. Afterwards this new SR image is used to generate the new set of LR images. If this new set of LR images is the same the earlier set, then the assumed SR image is the true SR image, otherwise the error image obtained from the difference between then LR images are back projected to the assumed SR image. This process is being repeated till no error image is left. IBP can be mathematically represented as

$$f^{(n+1)}(x) = f^n(x) + \sum_y \left(g_k(y) - g_k^{(n)}(y)\right) \times H^{BP} \quad (8)$$

where $f^n$ is estimated SR image after n iteration, $g_k^{(n)}$ are calculated LR images from the imaging model of $f^n$ after n iteration and $H^{BP}$ is the back projection kernel.

The second part deals with the image registration. Irani and Peleg [22] studied on the image registration method in order to achieve a more general motion model. Since previous works done in [22] limits its efficiency due to taking account on remote sensing images.

The proposed SR technique is benefitting from bicubic interpolation and IBP registration technique. The proposed SR technique is also compared with conventional bicubic interpolation, and wavelet zero padding (WZP) as well as the state-of-the-art technique proposed by Irani and Peleg [22].

## II. THE PROPOSED RESOLUTION ENHANCEMENT TECHNIQUE

The proposed SR technique is benefiting from both interpolation and IBP registration. First the four LR images obtained by taking some sequential images or artificially generated by using blurring filter, are being interpolated by using bicubic interpolation. The interpolated images are not sharp and due to the smoothing caused by interpolation a sharpening is required. Hence these four interpolated images are being used as input to the IBP registration technique. The output HR image is being obtained after the IBP stage. Fig. 1 is showing the block diagram of the proposed SR technique.

The experimental results given in the next section are showing the superiority of the proposed technique over the conventional and state-of-the-art techniques.

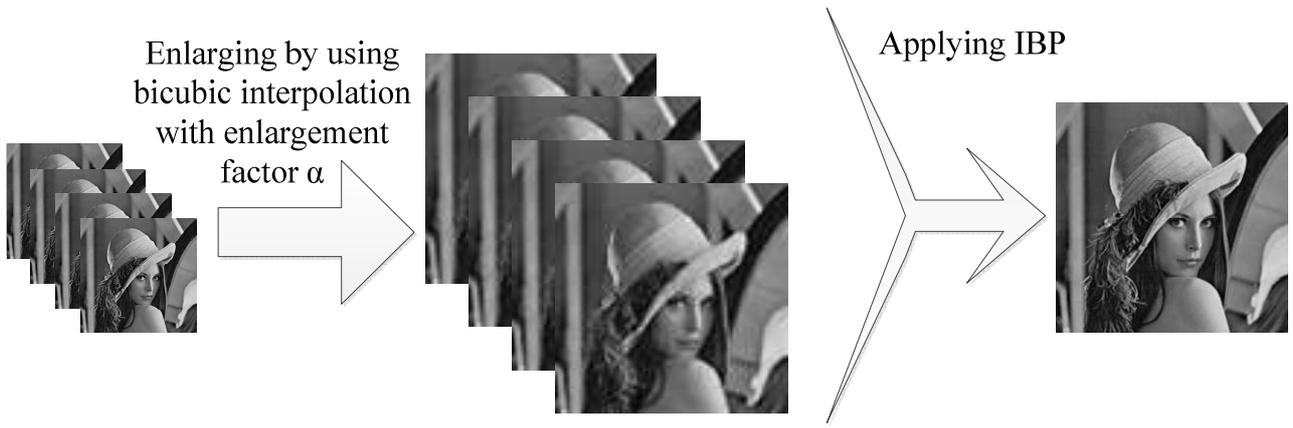

**Fig 1:** The block diagram of the proposed resolution enhancement technique.

## III. EXPERIMENTAL RESULTS AND DISCUSSION

For comparison purposes, the proposed SR techniques as well as other conventional and the state-of-the-art techniques are being tested on several well-known images namely, Lena, Elaine, Pepper, and Baboon. Table 1 shows the PSNR values in dB for Bicubic interpolation, WZP, Irani and Peleg, and the proposed SR technique of the aforementioned images. The LR images are 128x128 and the HR images are 256x256.

**Table 1:** The PSNR values (dB) for resolution enhancement of different images by using several SR techniques

| Technique | PSNR Value in dB | | | |
|---|---|---|---|---|
| | Lena | Baboon | Peppers | Elaine |
| Bicubic Interpolation | 18.60 | 19.54 | 21.07 | 22.18 |
| Irani and Peleg [22] | 24.47 | 20.40 | 25.03 | 27.25 |
| WZP (db.9/7) | 23.59 | 17.07 | 22.11 | 26.87 |
| Proposed technique | 24.73 | 20.81 | 25.91 | 27.97 |

For further comparison, Table 2 is prepared which shows the performance of the proposed SR technique compared to the other techniques from structural similarity index (SSIM) point of view [28].

**Table 2:** The SSIM for resolution enhancement of different images by using several SR techniques

| Technique | SSIM | | | |
|---|---|---|---|---|
| | Lena | Baboon | Peppers | Elaine |
| Bicubic Interpolation | 0.5827 | 0.4037 | 0.6964 | 0.7186 |
| Irani and Peleg [22] | 0.8284 | 0.5129 | 0.8324 | 0.8925 |
| WZP (db.9/7) | 0.8111 | 0.6250 | 0.8170 | 0.8764 |
| Proposed technique | 0.8579 | 0.5827 | 0.8604 | 0.9138 |

Fig. 2 is showing the visual comparison between bicubic interpolation, Irani and Peleg SR technique and the proposed SR technique for Lena image.

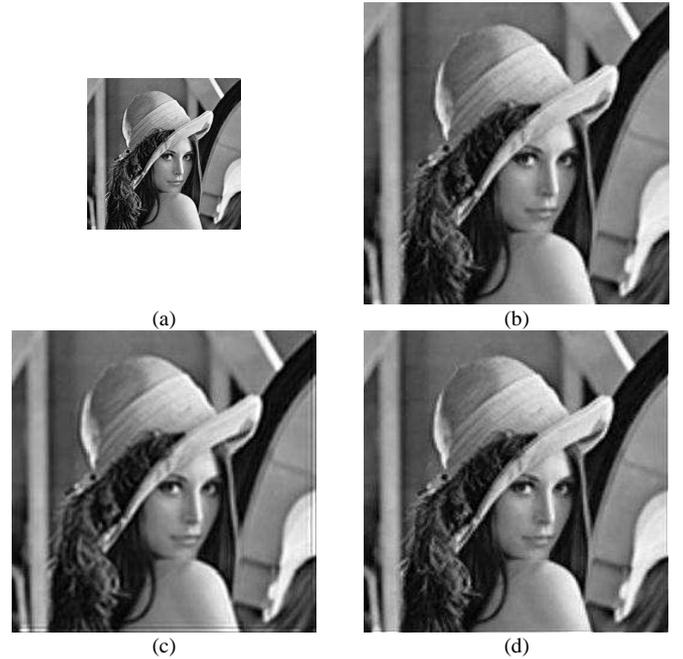

**Fig. 2:** The visual comparison between: (a) original low resolution Lena image (128x128) and the super resolved image (256x256) by using (b) bicubic interpolation, (c) Irani and Pelege SR technique, and (d) the proposed technique.

## IV. CONCLUSION

In this research work a new SR technique based on the interpolation followed by registering them using IBP was proposed. First four LR images were interpolated and then registered by using IBP in order to generate a sharper HR image. The proposed technique was tested on Lena, Elaine, Pepper, and Baboon and compared by conventional and the state-of-the-art techniques by means of PSNR and SSIM. Quantitative and qualitative results showed the superiority of the proposed technique over the other SR techniques.